\documentclass[
twocolumn,
]{ceurart}

\usepackage{gensymb}

\begin{document}

\copyrightyear{2021}
\copyrightclause{Copyright for this paper by its authors.
  Use permitted under Creative Commons License Attribution 4.0
  International (CC BY 4.0).}

\conference{1st Workshop on Complex Data Challenges in Earth Observation (CDCEO) 2021,
  November 01, 2021}

\title{Spatiotemporal Weather Data Predictions with Shortcut Recurrent-Convolutional Networks: A Solution for the Weather4cast challenge}

\author[1]{Jussi Leinonen}[%
orcid=0000-0002-6560-6316,
email=jussi.leinonen@meteoswiss.ch,
url=https://jleinonen.github.io/,
]
\address[1]{Federal Office of Meteorology and Climatology MeteoSwiss, Locarno-Monti, Switzerland}

\begin{abstract}
This paper presents the neural network model that was used by the author in the Weather4cast 2021 Challenge Stage 1, where the objective was to predict the time evolution of satellite-based weather data images. The network is based on an encoder-forecaster architecture making use of gated recurrent units (GRU), residual blocks and a contracting/expanding architecture with shortcuts similar to U-Net. A GRU variant utilizing residual blocks in place of convolutions is also introduced. Example predictions and evaluation metrics for the model are presented. These demonstrate that the model can retain sharp features of the input for the first predictions, while the later predictions become more blurred to reflect the increasing uncertainty.
\end{abstract}


\maketitle

\section{Introduction}

Weather forecasting requires the prediction of spatiotemporal changes in atmospheric fields. Most commonly, this prediction is carried out using numerical weather prediction (NWP, e.g. \cite{Bauer2015NWP}), which models the evolution of the atmosphere by computationally evaluating approximate physical laws, such as forward-integrating the partial differential equations of fluid dynamics. 

Although the NWP approach to forecasting is firmly grounded in first principles, NWP models are computationally expensive: state-of-the-art models are run on powerful supercomputers and yet take a significant amount of time to complete a model run, which limits their usefulness in short-term predictions. These issues have motivated considerable research on using artificial intelligence (AI) and machine learning (ML) methods for weather prediction, particularly after the prominent progress in convolutional neural networks (ConvNets) in the last decade. ConvNets can learn to recognize complex spatial structures, which make them well-suited for modeling atmospheric fields. They have been used as components of weather models, for example, to speed up computationally expensive modules of NWP codes. Recurrent neural networks have also been used to model the time evolution of atmospheric fields. Numerous applications and potential uses of ML/AI in weather in climate sciences have been discussed by, e.g., \cite{McGovern2017AIWeather,Reichstein2019DLProcess,Huntingford2019MLClimate,Haupt2021MLPostprocessing}. Thus, there is considerable interest in further developing ML model architectures for these applications.

To support the development of ML/AI for weather prediction, the Weather4cast 2021 Challenge had the objective to predict the time evolution of atmospheric satellite data products using machine-learning methods. This paper presents the ML solution submitted by the author.

\section{Challenge Data and Objectives}
The Weather4cast 2021 Challenge Stage 1 presented the competitors with a data problem of predicting the time evolution of satellite-based measurements of four different atmospheric fields: \textit{temperature}, representing either the surface temperature (where no cloud is present) or the cloud top temperature (in cloudy areas); convective rainfall rate (\textit{crr\_intensity}); probability of tropopause folding (\textit{asii\_turb\_trop\_prob}); and cloud mask (\textit{cma}).
These are data products created by the Nowcasting Satellite Application Facility (NWC SAF) processing from satellite data from the EUMETSAT-operated MeteoSat satellites. The products were given for six regions (numbered R1--R6) in Europe, Northern Africa and the Middle East, each $256 \times 256$ pixels of approximately $4\ \mathrm{km} \times 4\ \mathrm{km}$ each. A few other variables also were available such as cloud type, as well as static data with the elevation of the terrain, the latitude and the longitude.

The goal of the challenge was to predict, in each area, the next $32$ frames of a sequence given the previous $4$ frames, with frames separated by $15\ \mathrm{min}$. Training and validation datasets, with complete $4+32$-frame samples, as well as a test dataset with only the $4$-frame inputs, were provided for regions R1--R3, constituting the ``Core'' competition. Meanwhile, R4--R6 only had test data available, meaning that they had to be evaluated using models trained on R1--R3; this was called the ``Transfer Learning'' competition. Furthermore, all regions had a set of ``held-out'' data which were made available only during the final week of the competition; the final results were based on the performance with these data.

The performance of the models was evaluated using the mean-square error (MSE) for each variable. However, some adjustments were made to the MSE to account for the particular needs of each variable, except for \textit{crr\_intensity}. First, the loss for \textit{temperature} was modified to account for varying amounts of missing data in each region. Second, \textit{asii\_turb\_trop\_prob} is a probabilistic variable and the output of the model was passed through a truncated and normalized logit transform before the evaluation of the MSE. Third, although \textit{cma} is technically evaluated using the MSE, the variable in the output data file is required to be quantized such that the value is either $0$ or $1$; therefore, model output values $<0.5$ are rounded to $0$ and outputs $\geq 0.5$ are rounded to $1$ before evaluation. The details of the metrics can be found in \cite{IARAI2021Metrics}.

\section{Solution}

\subsection{Models} \label{sect:models}

\begin{figure*}[b]
  \centering
  \includegraphics[width=0.8\linewidth]{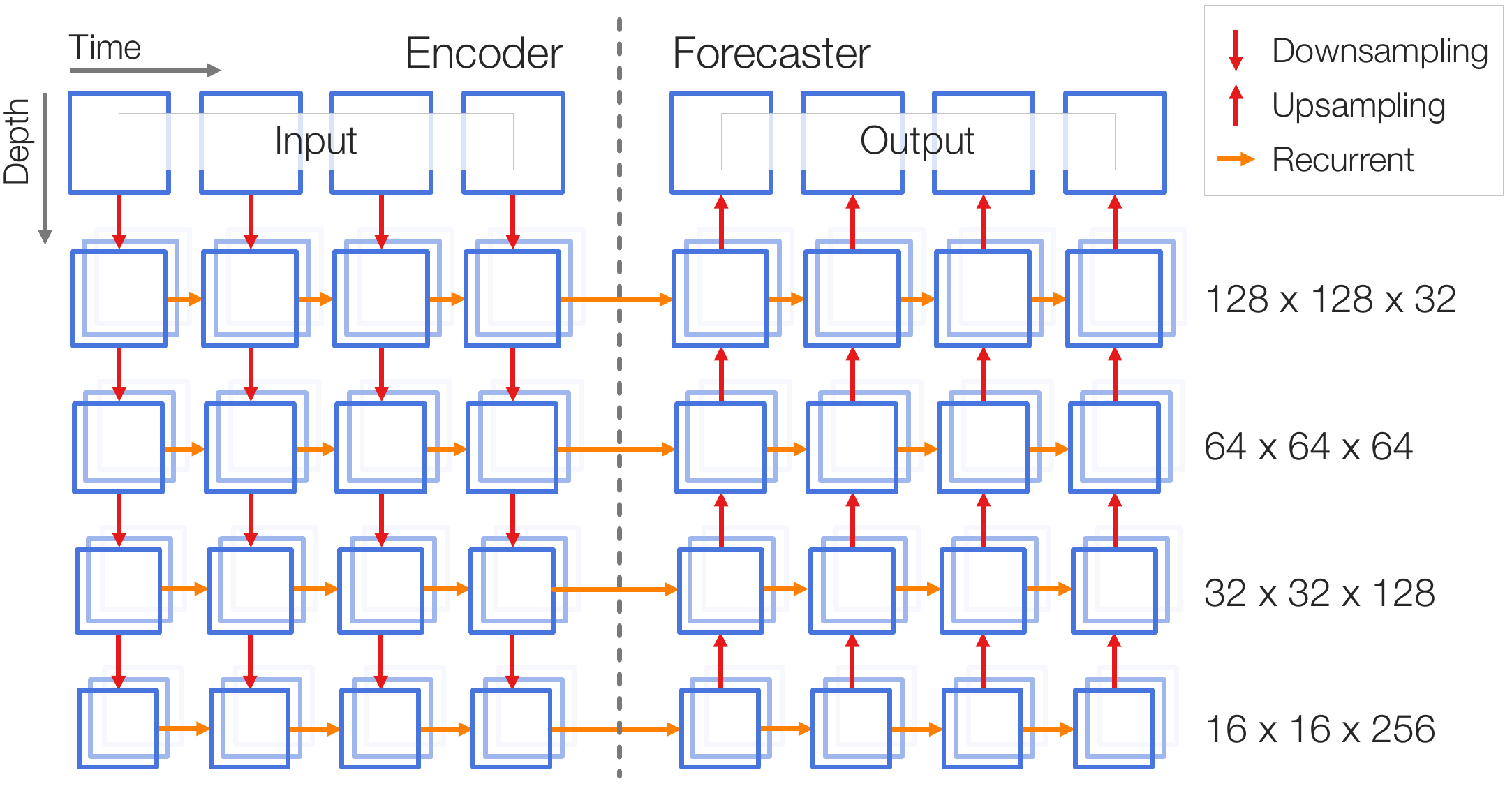}
  \caption{Illustration of the network architecture.} \label{fig:w4cnetwork}
\end{figure*}

The model presented here is a neural network combining recurrent-convolutional layers and shortcut connections in an encoder-forecaster architecture. The architecture is presented in Fig.~\ref{fig:w4cnetwork}. It is based on that developed in \cite{Shi2017DLPrecip} for precipitation nowcasting and adopted by \cite{Franch2020Nowcast}, as well as similar to that of \cite{Ravuri2021Nowcast}, with some differences that are described below.

The encoder section consists of four recurrent downsampling stages. Each stage first passes the sequence through a residual block \cite{He2016ResNet}, with each frame processed using the same convolutional filters. A strided convolution in the residual block is used to downsample the input by a factor of 2. Then, the sequence is processed by a gated recurrent unit (GRU) layer \cite{Cho2014GRU}; a tensor of zeros is passed as the initial state of the GRU. The number of channels in the convolutions is increased with increasing depth in the encoder.

The forecaster section is approximately a mirror image of the encoder section. Each stage consists of a GRU layer which is followed by bilinear upsampling and a residual block. A shortcut similar to U-Net \cite{Ronneberger2015UNet} is utilized: The final state of each GRU in the encoder is passed through a convolution and then used as the initial state of the GRU of corresponding depth in the forecaster. This allows the high-resolution features of the recent frames to be passed through, preventing the first predictions from being blurry. A final projection and a sigmoid activation produce the output as a single variable constrained between $0$ and $1$.

The main difference of the architecture presented here to that of \cite{Shi2017DLPrecip} is that the use of Trajectory GRU (TrajGRU) is rejected as TrajGRU was found to cause training instability. Two variants are considered instead. The first utilizes the Convolutional GRU (ConvGRU) layer adopted by e.g. \cite{Ravuri2021Nowcast,Tian2020GRUNowcasting,Leinonen2020Downscaling}. In the second variant, the convolution in the ConvGRU is replaced by a residual block modified to be used for this purpose. The use of the residual block increases the depth of the operations in the GRU and is thus expected to allow it to better process nonlinear transformations and also to increase the distance at which pixels can influence each other at each step of the ConvGRU. The latter effect may recover some of the advantages of TrajGRU over ConvGRU that \cite{Shi2017DLPrecip} found. The author is unaware of previously published instances of a residual layer being used in place of the convolution in GRU. In this paper, this variant is called ``ResGRU'', although the same abbreviation was used for a different combination of GRUs and residual connections in  \cite{Gao2020ResGRU}.

The models were implemented using TensorFlow/Keras \cite{Chollet2015Keras} version~2.4. The source code and the pre-trained models can be found through the links in Appendix~\ref{sect:code}.

\subsection{Training}

Since the scores for the target variables were evaluated independently from each other, a separate instance of the model was trained for each target variable, but using all variables as inputs for each model. The models were trained on the training dataset of R1--R3 such that every available gapless sequence of $36$ frames was used for training, resulting in $72192$ different sequences (albeit with considerable overlap). The training was performed with combined data from all regions R1--R3 in order to increase the training dataset size and improve the ability of the model to generalize; specializing the model to single regions was not attempted. The static data (latitude, longitude and elevation) were also used for training. Data augmentation by random rotation in $90\degree$ increments as well as random top-down and left-right mirroring was used to further increase the effective number of training samples. The model for \textit{asii\_turb\_trop\_prob} was trained using a custom logit loss corresponding to the metric specified in \cite{IARAI2021Metrics}, while the other variables were trained using the standard MSE loss. 
%It was speculated that the quantization of \textit{cma} might mean that it would be better predicted by another loss, and the Dice loss was attempted for this purpose, but this led to inferior results compared to MSE.
The Adam optimizer \cite{Kingma2014Adam} was used to train the models with a batch size of $32$.

The progress of the training was evaluated using the provided validation dataset for R1--R3. After each training epoch, the evaluation metric was computed on the validation set and then:
\begin{enumerate}
    \item If the metric improved upon the best evaluation result, the model weights were saved.
    \item If the metric had not improved in $3$ epochs, the learning rate was reduced by a factor of $5$.
    \item If the metric had not improved in $10$ epochs, the training was stopped early. \label{item:early_stop}
\end{enumerate}
In practice, condition~\ref{item:early_stop} was never activated as the model continued to achieve marginal gains on the validation data at least every few epochs until the maximum training time of $12\ \mathrm{h}$ or $24\ \mathrm{h}$ (depending on the training run) was reached. This suggests that the model did not suffer significantly from overfitting, which typically causes the validation loss to start increasing even as the training loss keeps decreasing. This is perhaps due to the relatively modest number of weights in the models by standards of modern ConvNets, approximately $12.1$ million weights in the ConvGRU variant and $18.6$ million in the ResGRU variant. The loss over the validation set was used as the metric for each variable except \textit{cma} for which a rounded MSE that takes the ${0,1}$ quantization into account was used.

A parallel setup of eight Nvidia Tesla V100 GPUs was used to train the models. Training for one epoch took approximately 20 minutes with this hardware. The eight parallel GPUs only provided a speedup of a factor of approximately $3$ compared to training on a single GPU, suggesting that single-GPU training of the models should be feasible, although the batch size would likely have to be reduced as the models require rather large amounts of GPU memory.

\section{Results}

Both the ConvGRU and ResGRU variants of the model were trained for each target variable. The evaluation results for the validation dataset are shown in Table~\ref{table:validation}. Comparisons to TrajGRU were found impractical as the models using TrajGRU would not converge properly due to the training instability mentioned in Sect.~\ref{sect:models}.
\begin{table*}
  \caption{Evaluation metrics for the validation dataset.}
  \label{table:validation}
  \begin{tabular}{ccccc}
    \toprule
    & {\it temperature} & {\it crr\_intensity} & {\it asii\_turb\_trop\_prob} & {\it cma}\\
    \hline
    ConvGRU & $0.004564$ & $\mathbf{0.0001259}$ & $0.002250$ & $0.1393$\\
    ResGRU & $\mathbf{0.004356}$ & $0.0001278$ & $\mathbf{0.002161}$ & $\mathbf{0.1376}$\\
  \bottomrule
  \end{tabular}
\end{table*}

Based on the evaluation results, three submissions were made to the final leaderboards of Weather4cast Stage 1: one using the ConvGRU variant for all variables (codenamed V4c), another using ResGRU (V4rc), and a third using the best model for each variable based on the validation metrics (V4pc). It was indeed this last model that produced the best results also on the leaderboards for both the Core and Transfer Learning competitions, as shown in Table~\ref{table:heldout}.
\begin{table}
  \caption{Evaluation metrics for the held-out test dataset, as computed by the Weather4cast website (\url{https://www.iarai.ac.at/weather4cast/}).}
  \label{table:heldout}
  \begin{tabular}{ccc}
    \toprule
    & Core & Transfer learning\\
    \hline
    ConvGRU & $0.5051$ & $0.4658$ \\
    ResGRU & $0.5014$ & $0.4626$ \\
    Best combination & $\mathbf{0.4987}$ & $\mathbf{0.4607}$\\
  \bottomrule
  \end{tabular}
\end{table}

\begin{figure*}
  \centering
  \includegraphics[width=\linewidth]{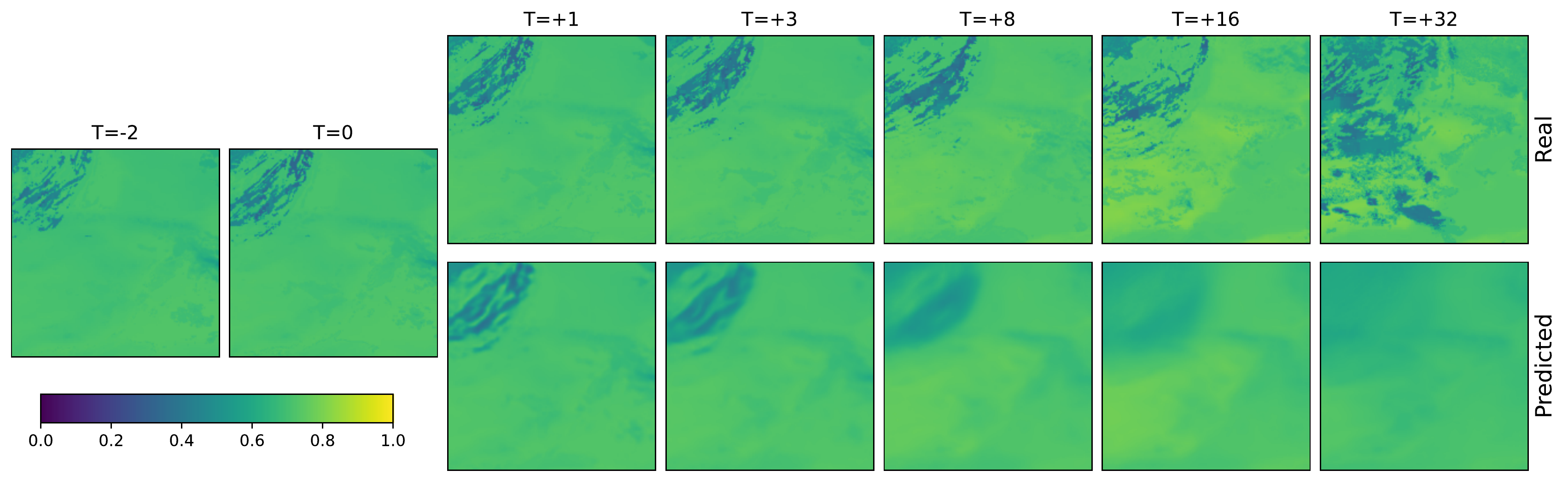}
  \caption{An example of predictions for the \textit{temperature} variable. The frames on the left correspond to past temperature, while the frames on the right show the real future temperature (top row) and the predicted temperature (bottom row). The $T$ coordinate refers to the index of the frame in the sequence, with $T=0$ represents the last input data point and $T=1$ the first prediction. The model output normalized to the range $(0,1)$ is shown.} \label{fig:prediction_temp}
\end{figure*}

\begin{figure*}
  \centering
  \includegraphics[width=\linewidth]{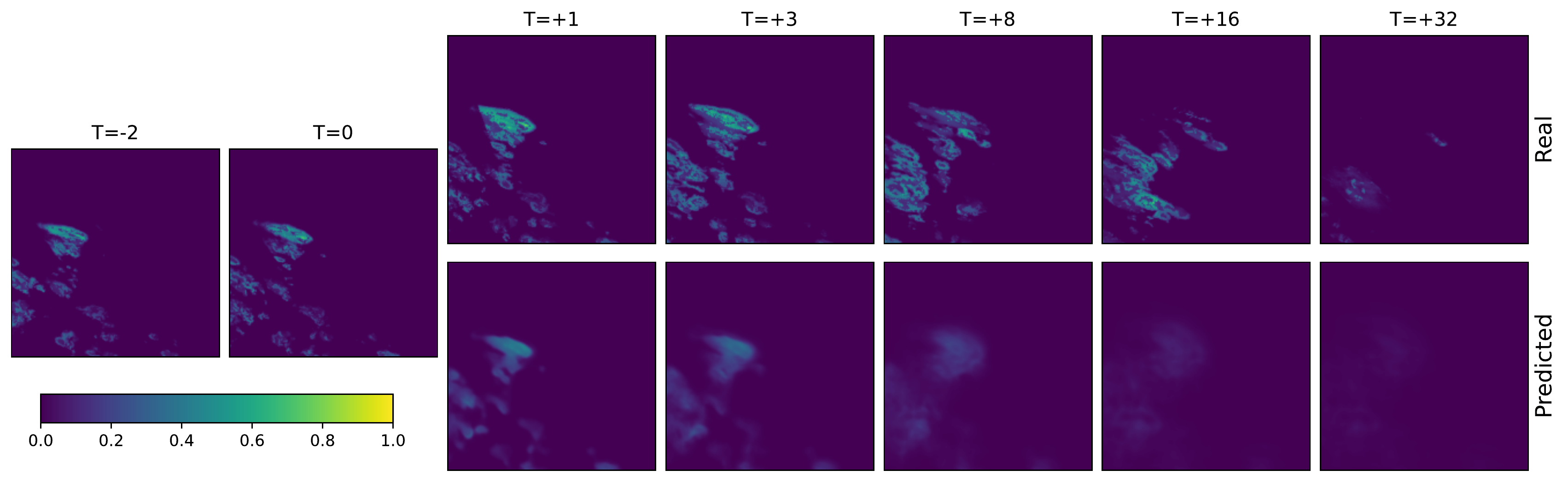}
  \caption{As Fig.~\ref{fig:prediction_temp}, but for \textit{crr\_intensity}. A different case is shown as the case of Fig.~\ref{fig:prediction_temp} does not contain precipitation.} \label{fig:prediction_crr}
\end{figure*}

\begin{figure*}
  \centering
  \includegraphics[width=\linewidth]{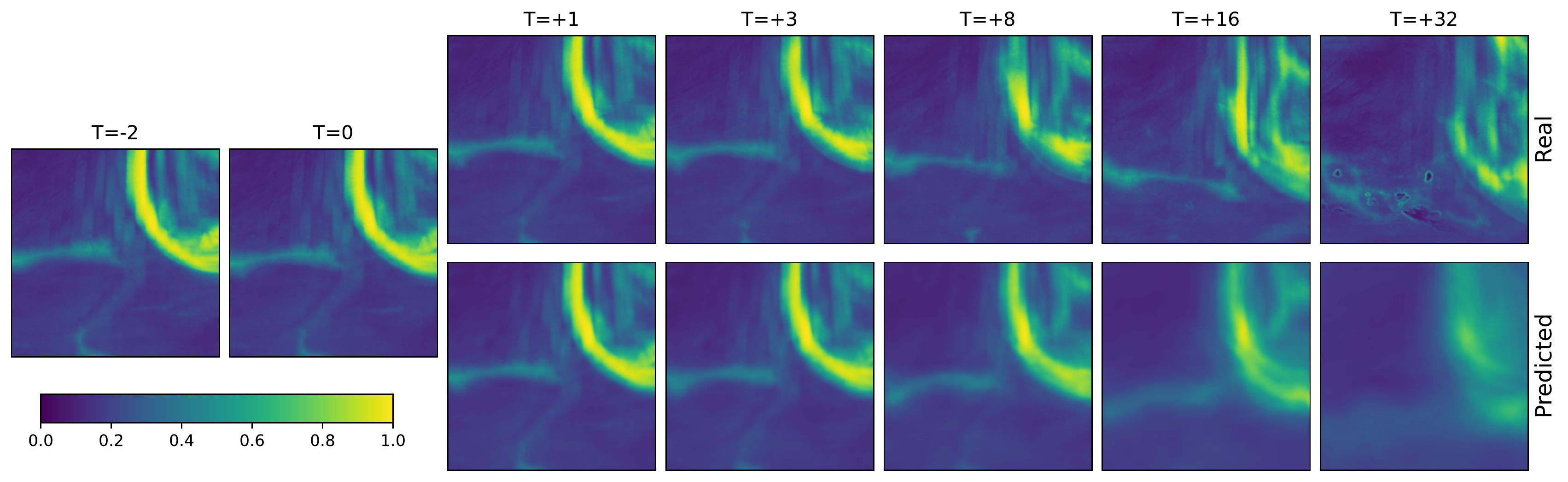}
  \caption{As Fig.~\ref{fig:prediction_temp}, but for \textit{asii\_turb\_trop\_prob}.} \label{fig:prediction_attp}
\end{figure*}

\begin{figure*}
  \centering
  \includegraphics[width=\linewidth]{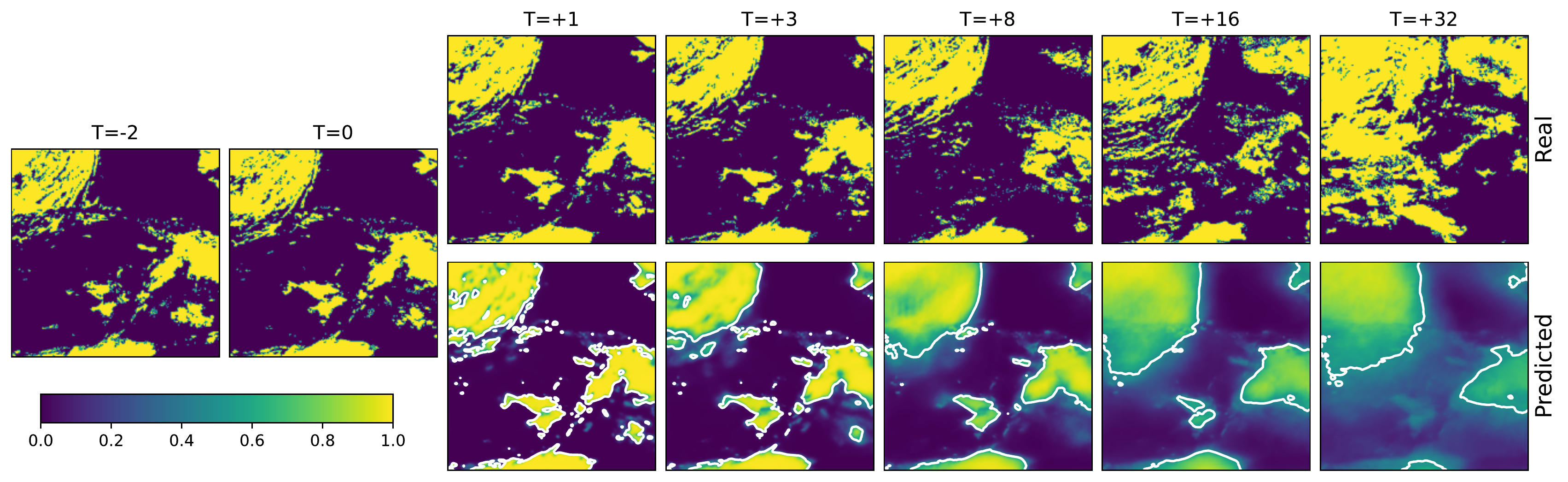}
  \caption{As Fig.~\ref{fig:prediction_temp}, but for \textit{cma}. The white contours in the predictions indicate $0.5$, the threshold of the cloud mask in the output.} \label{fig:prediction_cma}
\end{figure*}

Figures~\ref{fig:prediction_temp}--\ref{fig:prediction_cma} show examples of the predictions using the validation dataset. These are all shown for the same scene except for Fig.~\ref{fig:prediction_crr}, where a different scene was chosen because the one used for the others did not contain precipitation. It is clear that the predictions start relatively sharp and get blurrier over time, reflecting the increasing uncertainty. The blurriness is likely exacerbated by the use of the MSE metric, specified in the data challenge, which is prone to regression to the mean. Especially in Fig.~\ref{fig:prediction_attp}, one can also see that the model can predict the motion of features in the images.

\section{Conclusions}

The model presented here reached the top of the final leaderboards in both the Core and the Transfer Learning categories of the Weather4cast 2021 Challenge Stage 1. It is a versatile solution to the problem of predicting the evolution of atmospheric fields, producing sharp predictions for the near term and increasing the uncertainty for longer lead times. The architecture can be easily adapted to other tasks such as probabilistic predictions or outputs that are different from the inputs. Further research is needed to handle, for instance, different spatial and temporal resolutions of inputs and data available for future time steps.

%\begin{table*}
%  \caption{Frequency of Special Characters}
%  \label{tab:freq}
%  \begin{tabular}{ccl}
%    \toprule
%    Non-English or Math&Frequency&Comments\\
%    \midrule
%    \O & 1 in 1,000& For Swedish names\\
%    $\pi$ & 1 in 5& Common in math\\
%    \$ & 4 in 5 & Used in business\\
%    $\Psi^2_1$ & 1 in 40,000& Unexplained usage\\
%  \bottomrule
%\end{tabular}
%\end{table*}

%\begin{figure}
%  \centering
%  \includegraphics[width=\linewidth]{sample-franklin}
%  \caption{1907 Franklin Model D roadster. Photograph by Harris \&
%    Ewing, Inc. [Public domain], via Wikimedia
%    Commons. (\url{https://goo.gl/VLCRBB}).}
%\end{figure}

\begin{acknowledgments}
This project benefited from parallel development in the fellowship ``Seamless Artificially Intelligent Thunderstorm Nowcasts'' from the European Organisation for the Exploitation of Meteorological Satellites (EUMETSAT). The hosting institution of this fellowship is MeteoSwiss in Switzerland. The author thanks U. Hamann and A. Rigazzi for discussions regarding the model and training.
\end{acknowledgments}

%%
%% Define the bibliography file to be used
\bibliography{weather4cast}

\begin{thebibliography}{18}
\expandafter\ifx\csname natexlab\endcsname\relax\def\natexlab#1{#1}\fi
\providecommand{\url}[1]{\texttt{#1}}
\providecommand{\href}[2]{#2}
\providecommand{\path}[1]{#1}
\providecommand{\DOIprefix}{doi:}
\providecommand{\ArXivprefix}{arXiv:}
\providecommand{\URLprefix}{URL: }
\providecommand{\Pubmedprefix}{pmid:}
\providecommand{\doi}[1]{\href{http://dx.doi.org/#1}{\path{#1}}}
\providecommand{\Pubmed}[1]{\href{pmid:#1}{\path{#1}}}
\providecommand{\bibinfo}[2]{#2}
\ifx\xfnm\relax \def\xfnm[#1]{\unskip,\space#1}\fi
%Type = Article
\bibitem[{Bauer et~al.(2015)Bauer, Thorpe, and Brunet}]{Bauer2015NWP}
\bibinfo{author}{P.~Bauer}, \bibinfo{author}{A.~Thorpe},
  \bibinfo{author}{G.~Brunet},
\newblock \bibinfo{title}{The quiet revolution of numerical weather
  prediction},
\newblock \bibinfo{journal}{Nature} \bibinfo{volume}{525}
  (\bibinfo{year}{2015}) \bibinfo{pages}{47--55}.
  \DOIprefix\doi{doi.org/10.1038/nature14956}.
%Type = Article
\bibitem[{McGovern et~al.(2017)McGovern, Elmore, Gagne, Haupt, Karstens,
  Lagerquist, Smith, and Williams}]{McGovern2017AIWeather}
\bibinfo{author}{A.~McGovern}, \bibinfo{author}{K.~L. Elmore},
  \bibinfo{author}{D.~J. Gagne, II}, \bibinfo{author}{S.~E. Haupt},
  \bibinfo{author}{C.~D. Karstens}, \bibinfo{author}{R.~Lagerquist},
  \bibinfo{author}{T.~Smith}, \bibinfo{author}{J.~K. Williams},
\newblock \bibinfo{title}{Using artificial intelligence to improve real-time
  decision-making for high-impact weather},
\newblock \bibinfo{journal}{Bull. Amer. Meteor. Soc.} \bibinfo{volume}{98}
  (\bibinfo{year}{2017}) \bibinfo{pages}{2073--2090}.
  \DOIprefix\doi{10.1175/BAMS-D-16-0123.1}.
%Type = Article
\bibitem[{Reichstein et~al.(2019)Reichstein, Camps-Valls, Stevens, Jung,
  Denzler, Carvalhais, and Prabhat}]{Reichstein2019DLProcess}
\bibinfo{author}{M.~Reichstein}, \bibinfo{author}{G.~Camps-Valls},
  \bibinfo{author}{B.~Stevens}, \bibinfo{author}{M.~Jung},
  \bibinfo{author}{J.~Denzler}, \bibinfo{author}{N.~Carvalhais},
  \bibinfo{author}{Prabhat},
\newblock \bibinfo{title}{Deep learning and process understanding for
  data-driven earth system science},
\newblock \bibinfo{journal}{Nature} \bibinfo{volume}{566}
  (\bibinfo{year}{2019}) \bibinfo{pages}{195--204}.
  \DOIprefix\doi{10.1038/s41586-019-0912-1}.
%Type = Article
\bibitem[{Huntingford et~al.(2019)Huntingford, Jeffers, Bonsall, Christensen,
  Lees, and Yang}]{Huntingford2019MLClimate}
\bibinfo{author}{C.~Huntingford}, \bibinfo{author}{E.~S. Jeffers},
  \bibinfo{author}{M.~B. Bonsall}, \bibinfo{author}{H.~M. Christensen},
  \bibinfo{author}{T.~Lees}, \bibinfo{author}{H.~Yang},
\newblock \bibinfo{title}{Machine learning and artificial intelligence to aid
  climate change research and preparedness},
\newblock \bibinfo{journal}{Environmental Research Letters}
  \bibinfo{volume}{14} (\bibinfo{year}{2019}) \bibinfo{pages}{124007}.
  \DOIprefix\doi{10.1088/1748-9326/ab4e55}.
%Type = Article
\bibitem[{Haupt et~al.(2021)Haupt, Chapman, Adams, Kirkwood, Hosking, Robinson,
  Lerch, and Subramanian}]{Haupt2021MLPostprocessing}
\bibinfo{author}{S.~E. Haupt}, \bibinfo{author}{W.~Chapman},
  \bibinfo{author}{S.~V. Adams}, \bibinfo{author}{C.~Kirkwood},
  \bibinfo{author}{J.~S. Hosking}, \bibinfo{author}{N.~H. Robinson},
  \bibinfo{author}{S.~Lerch}, \bibinfo{author}{A.~C. Subramanian},
\newblock \bibinfo{title}{Towards implementing artificial intelligence
  post-processing in weather and climate: proposed actions from the oxford 2019
  workshop},
\newblock \bibinfo{journal}{Philos. Trans. R. Soc. London, Ser. A}
  \bibinfo{volume}{379} (\bibinfo{year}{2021}) \bibinfo{pages}{20200091}.
  \DOIprefix\doi{10.1098/rsta.2020.0091}.
%Type = Misc
\bibitem[{IARAI(2021)}]{IARAI2021Metrics}
\bibinfo{author}{IARAI}, \bibinfo{title}{Weather4cast 2021: Competition
  metrics}, \bibinfo{year}{2021}. \URLprefix
  \url{https://www.iarai.ac.at/weather4cast/wp-content/uploads/sites/3/2021/04/w4c.pdf}.
%Type = Inproceedings
\bibitem[{Shi et~al.(2017)Shi, Gao, Lausen, Wang, Yeung, Wong, and
  WOO}]{Shi2017DLPrecip}
\bibinfo{author}{X.~Shi}, \bibinfo{author}{Z.~Gao},
  \bibinfo{author}{L.~Lausen}, \bibinfo{author}{H.~Wang},
  \bibinfo{author}{D.-Y. Yeung}, \bibinfo{author}{W.-k. Wong},
  \bibinfo{author}{W.-c. WOO},
\newblock \bibinfo{title}{Deep learning for precipitation nowcasting: A
  benchmark and a new model},
\newblock in: \bibinfo{editor}{I.~Guyon}, \bibinfo{editor}{U.~V. Luxburg},
  \bibinfo{editor}{S.~Bengio}, \bibinfo{editor}{H.~Wallach},
  \bibinfo{editor}{R.~Fergus}, \bibinfo{editor}{S.~Vishwanathan},
  \bibinfo{editor}{R.~Garnett} (Eds.), \bibinfo{booktitle}{Advances in Neural
  Information Processing Systems}, volume~\bibinfo{volume}{30},
  \bibinfo{publisher}{Curran Associates, Inc.}, \bibinfo{year}{2017}.
  \URLprefix
  \url{https://proceedings.neurips.cc/paper/2017/file/a6db4ed04f1621a119799fd3d7545d3d-Paper.pdf}.
%Type = Article
\bibitem[{Franch et~al.(2020)Franch, Nerini, Pendesini, Coviello, Jurman, and
  Furlanello}]{Franch2020Nowcast}
\bibinfo{author}{G.~Franch}, \bibinfo{author}{D.~Nerini},
  \bibinfo{author}{M.~Pendesini}, \bibinfo{author}{L.~Coviello},
  \bibinfo{author}{G.~Jurman}, \bibinfo{author}{C.~Furlanello},
\newblock \bibinfo{title}{Precipitation nowcasting with orographic enhanced
  stacked generalization: Improving deep learning predictions on extreme
  events},
\newblock \bibinfo{journal}{Atmosphere} \bibinfo{volume}{11}
  (\bibinfo{year}{2020}). \DOIprefix\doi{10.3390/atmos11030267}.
%Type = Misc
\bibitem[{Ravuri et~al.(2021)Ravuri, Lenc, Willson, Kangin, Lam, Mirowski,
  Fitzsimons, Athanassiadou, Kashem, Madge, Prudden, Mandhane, Clark, Brock,
  Simonyan, Hadsell, Robinson, Clancy, Arribas, and
  Mohamed}]{Ravuri2021Nowcast}
\bibinfo{author}{S.~Ravuri}, \bibinfo{author}{K.~Lenc},
  \bibinfo{author}{M.~Willson}, \bibinfo{author}{D.~Kangin},
  \bibinfo{author}{R.~Lam}, \bibinfo{author}{P.~Mirowski},
  \bibinfo{author}{M.~Fitzsimons}, \bibinfo{author}{M.~Athanassiadou},
  \bibinfo{author}{S.~Kashem}, \bibinfo{author}{S.~Madge},
  \bibinfo{author}{R.~Prudden}, \bibinfo{author}{A.~Mandhane},
  \bibinfo{author}{A.~Clark}, \bibinfo{author}{A.~Brock},
  \bibinfo{author}{K.~Simonyan}, \bibinfo{author}{R.~Hadsell},
  \bibinfo{author}{N.~Robinson}, \bibinfo{author}{E.~Clancy},
  \bibinfo{author}{A.~Arribas}, \bibinfo{author}{S.~Mohamed},
  \bibinfo{title}{Skillful precipitation nowcasting using deep generative
  models of radar}, \bibinfo{year}{2021}.
  \href{http://arxiv.org/abs/2104.00954}{{\tt arXiv:2104.00954}}.
%Type = Inproceedings
\bibitem[{He et~al.(2016)He, Zhang, Ren, and Sun}]{He2016ResNet}
\bibinfo{author}{K.~He}, \bibinfo{author}{X.~Zhang}, \bibinfo{author}{S.~Ren},
  \bibinfo{author}{J.~Sun},
\newblock \bibinfo{title}{Deep residual learning for image recognition},
\newblock in: \bibinfo{booktitle}{The IEEE Conference on Computer Vision and
  Pattern Recognition (CVPR)}, \bibinfo{year}{2016}.
  \DOIprefix\doi{10.1109/CVPR.2016.90}.
%Type = Inproceedings
\bibitem[{Cho et~al.(2014)Cho, {van}~Merrienboer, Bahdanau, and
  Bengio}]{Cho2014GRU}
\bibinfo{author}{K.~Cho}, \bibinfo{author}{B.~{van}~Merrienboer},
  \bibinfo{author}{D.~Bahdanau}, \bibinfo{author}{Y.~Bengio},
\newblock \bibinfo{title}{On the properties of neural machine translation:
  Encoder-decoder approaches},
\newblock in: \bibinfo{booktitle}{Proceedings of SSST-8, Eighth Workshop on
  Syntax, Semantics and Structure in Statistical Translation},
  \bibinfo{year}{2014}, pp. \bibinfo{pages}{103--111}.
%Type = Inproceedings
\bibitem[{Ronneberger et~al.(2015)Ronneberger, Fischer, and
  Brox}]{Ronneberger2015UNet}
\bibinfo{author}{O.~Ronneberger}, \bibinfo{author}{P.~Fischer},
  \bibinfo{author}{T.~Brox},
\newblock \bibinfo{title}{U-net: Convolutional networks for biomedical image
  segmentation},
\newblock in: \bibinfo{editor}{N.~Navab}, \bibinfo{editor}{J.~Hornegger},
  \bibinfo{editor}{W.~M. Wells}, \bibinfo{editor}{A.~F. Frangi} (Eds.),
  \bibinfo{booktitle}{Medical Image Computing and Computer-Assisted
  Intervention -- MICCAI 2015}, \bibinfo{year}{2015}, pp.
  \bibinfo{pages}{234--241}. \DOIprefix\doi{10.1007/978-3-319-24574-4_28}.
%Type = Article
\bibitem[{Tian et~al.(2020)Tian, Li, Ye, Xie, and Li}]{Tian2020GRUNowcasting}
\bibinfo{author}{L.~Tian}, \bibinfo{author}{X.~Li}, \bibinfo{author}{Y.~Ye},
  \bibinfo{author}{P.~Xie}, \bibinfo{author}{Y.~Li},
\newblock \bibinfo{title}{A generative adversarial gated recurrent unit model
  for precipitation nowcasting} \bibinfo{volume}{17} (\bibinfo{year}{2020})
  \bibinfo{pages}{601--605}. \DOIprefix\doi{10.1109/LGRS.2019.2926776}.
%Type = Article
\bibitem[{Leinonen et~al.(2020)Leinonen, Nerini, and
  Berne}]{Leinonen2020Downscaling}
\bibinfo{author}{J.~Leinonen}, \bibinfo{author}{D.~Nerini},
  \bibinfo{author}{A.~Berne},
\newblock \bibinfo{title}{Stochastic super-resolution for downscaling
  time-evolving atmospheric fields with a generative adversarial network},
\newblock \bibinfo{journal}{IEEE Trans. Geosci. Remote Sens.}
  (\bibinfo{year}{2020}). \DOIprefix\doi{10.1109/TGRS.2020.3032790}.
%Type = Article
\bibitem[{Gao and Wai(2020)}]{Gao2020ResGRU}
\bibinfo{author}{W.~Gao}, \bibinfo{author}{R.-J. Wai},
\newblock \bibinfo{title}{A novel fault identification method for photovoltaic
  array via convolutional neural network and residual gated recurrent unit},
\newblock \bibinfo{journal}{IEEE Access} \bibinfo{volume}{8}
  (\bibinfo{year}{2020}) \bibinfo{pages}{159493--159510}.
  \DOIprefix\doi{10.1109/ACCESS.2020.3020296}.
%Type = Misc
\bibitem[{Chollet et~al.(2015)}]{Chollet2015Keras}
\bibinfo{author}{F.~Chollet}, et~al., \bibinfo{title}{Keras},
  \bibinfo{howpublished}{\url{https://keras.io}}, \bibinfo{year}{2015}.
%Type = Inproceedings
\bibitem[{Kingma and Ba(2014)}]{Kingma2014Adam}
\bibinfo{author}{D.~P. Kingma}, \bibinfo{author}{J.~Ba},
\newblock \bibinfo{title}{Adam: A method for stochastic optimization},
\newblock in: \bibinfo{booktitle}{{3rd International Conference for Learning
  Representations, San Diego, California, USA}}, \bibinfo{year}{2014}.
  \URLprefix \url{https://arxiv.org/abs/1412.6980}.
%Type = Misc
\bibitem[{Leinonen(2021)}]{Leinonen2021Weights}
\bibinfo{author}{J.~Leinonen}, \bibinfo{title}{{Model weights for a
  Weather4cast 2021 Challenge Stage 1 solution}}, \bibinfo{year}{2021}.
  \DOIprefix\doi{10.5281/zenodo.5101213}.

\end{thebibliography}

\appendix

\section{Online Resources} \label{sect:code}

The source code with instructions to replicate the results presented in this paper can be found at \url{https://github.com/jleinonen/weather4cast-stage1}. The model weights used in the challenge submissions can be downloaded at \cite{Leinonen2021Weights}.

\end{document}